%% file: main.tex
\documentclass[conference]{IEEEtran}
\IEEEoverridecommandlockouts
\usepackage{amsmath,amssymb,amsfonts}
\usepackage{algorithm}
\usepackage{hyperref}
\usepackage{comment}
\usepackage{caption}
\usepackage{amsmath}
\usepackage{algpseudocode}
\usepackage{graphicx}
\graphicspath{{./images/}}
\usepackage{textcomp}
\usepackage{xcolor}
\def\BibTeX{{\rm B\kern-.05em{\sc i\kern-.025em b}\kern-.08em
    T\kern-.1667em\lower.7ex\hbox{E}\kern-.125emX}}
\usepackage
[maxbibnames=99,
firstinits=true,
sorting=none,
backend=biber]
{biblatex}

\hypersetup{
    colorlinks=true,
    linkcolor=black,
    filecolor=black,
    citecolor=black,
    urlcolor=blue,
    }
\begin{document}

\title{Computer-Vision Based Real Time Waypoint Generation for Autonomous Vineyard Navigation with Quadruped Robots\\
}

\author{\IEEEauthorblockN{Lee Milburn}
\IEEEauthorblockA{\textit{RIVeR Lab} \\
\textit{IIT \& NEU}\\
Boston, Massachusetts  \\
Milburn.l@northeastern.edu}

\and
\IEEEauthorblockN{Juan Gamba}
\IEEEauthorblockA{\textit{Dynamic Legged Systems Lab} \\
\textit{Istituto Italiano di Tecnologia}\\
Genova, Italy \\
Juan.Gamba@iit.it}
\and
\IEEEauthorblockN{Miguel Fernandes}
\IEEEauthorblockA{\textit{Advanced Robotics Lab} \\
\textit{IIT \& UniGe}\\
Genova, Italy \\
Miguel.Ferreira@iit.it}
\and
\IEEEauthorblockN{Claudio Semini}
\IEEEauthorblockA{\textit{Dynamic Legged Systems Lab} \\
\textit{Istituto Italiano di Tecnologia}\\
Genova, Italy \\
Claudio.Semini@iit.it}
}

\IEEEoverridecommandlockouts
\IEEEpubid{\makebox[\columnwidth]{979-8-3503-0121-2/23/\$31.00~\copyright2023 IEEE \hfill}
\hspace{\columnsep}\makebox[\columnwidth]{ }}
\maketitle

\begin{abstract}
The VINUM project seeks to address the shortage of skilled labor in modern vineyards by introducing a cutting-edge mobile robotic solution. Leveraging the capabilities of the quadruped robot, HyQReal, this system, equipped with arm and vision sensors, offers autonomous navigation and winter pruning of grapevines reducing the need for human intervention.
At the heart of this approach lies an architecture that empowers the robot to easily navigate vineyards, identify grapevines with unparalleled accuracy, and approach them for pruning with precision. A state machine drives the process, deftly switching between various stages to ensure seamless and efficient task completion.
The system's performance was assessed through experimentation, focusing on waypoint precision and optimizing the robot's workspace for single-plant operations. Results indicate that the architecture is highly reliable, with a mean error of 21.5cm and a standard deviation of 17.6cm for HyQReal. However, improvements in grapevine detection accuracy are necessary for optimal performance.
This work is based on a computer-vision-based navigation method for quadruped robots in vineyards, opening up new possibilities for selective task automation. The system's architecture works well in ideal weather conditions, generating and arriving at precise waypoints that maximize the attached robotic arm's workspace. This work is an extension of our
short paper presented at the Italian Conference on Robotics and
Intelligent Machines (I-RIM), 2022 \cite{irim}.
\end{abstract}


\section{Introduction}
\input{sections/1_introduction}

\section{State of the Art}
\input{sections/state_of_the_art.tex}

\section{Navigation Architecture}
\input{sections/2_navigation_architecture}

\section{Experiments}
\input{sections/4_experiment}

\section{Conclusion}
\input{sections/conclusion.tex}

\section{Acknowledgments}
\input{sections/7_acknowledgments}

\printbibliography
\end{document}

%% file: sections/1_introduction.tex
\begin{figure}[h]
\centering
\includegraphics[height=3.5cm, width=0.40\textwidth]{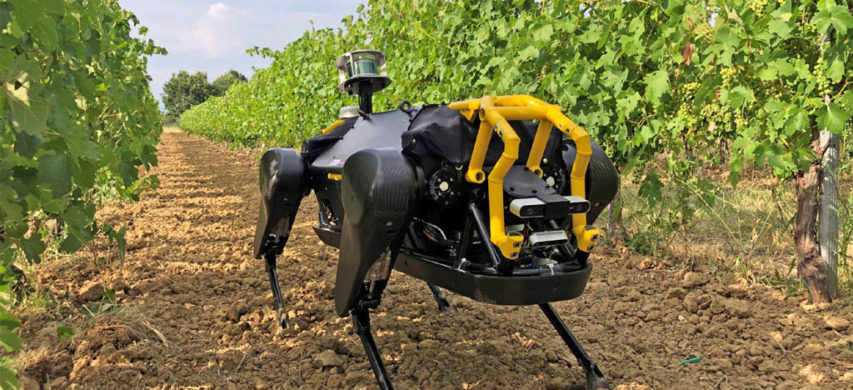}
\captionsetup{justification=centering}
\caption{IIT's HyQReal quadruped robot in a Vineyard close to Piacenza, Italy.}
\label{fig:Vinum}
\end{figure}

The wine industry has been experiencing a significant labor shortage in vineyards worldwide\footnote[1]{https://www.winemag.com/2021/12/07/wine-industry-labor-supply/}. Vineyards heavily rely on seasonal workers, including international workforces, for their labor needs. Consequently, they have been exploring the potential of robotic automation of seasonal work to address the labor shortage.
 
The VINUM\footnote[2]{https://vinum-robot.eu/} project uses IIT's HyQReal quadruped robot to do the winter pruning of grapevines autonomously with the Kivoa Gen3 robotic arm \cite{miguel}\cite{Guadagna2021}; see Fig. \ref{fig:Vinum}, \cite{hyqreal}. Quadruped robots can navigate obstacles and adjust their movements accordingly, making them well-suited for specific vineyards representing a complex and dynamic environment.  This improved maneuverability in conjunction with a robotic pruning arm leads to a more efficient pruning process and reduces the need for manual labor. Ultimately, these benefits can contribute to improved growth and yield of grapevines. For example, HyQReal can move laterally as well as forward and backward. This extra movement allows for more mobility when searching for grapevines and more freedom of approach when moving to detected grapevines \cite{BISWAL20212017}.

\begin{figure}[h]
\centering
\includegraphics[height=3.5cm, width=0.40\textwidth]{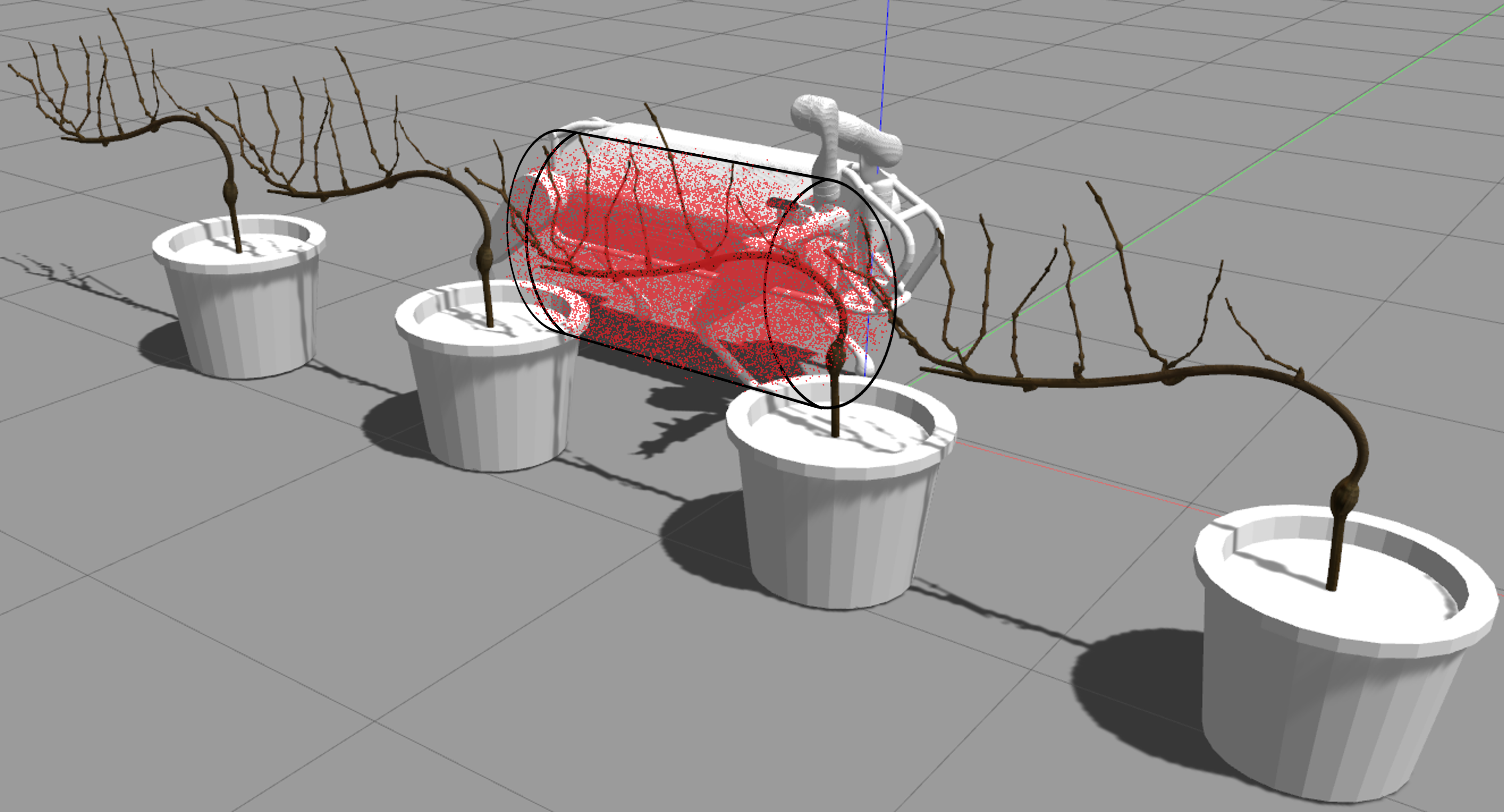}
\captionsetup{justification=centering}
\caption{Kinematic workspace (in red) of the Kinova Gen3 manipulator arm mounted on HyQReal.}
\label{fig:Workspace}
\end{figure}

To autonomously winter-prune grapevines, the VINUM robot has to reliably and autonomously navigate vineyards, arriving at each grapevine that needs winter pruning. To reach each grapevine for pruning, the navigation stack has to autonomously generate waypoints relative to the grapevine which are ideal for the robotic workspace of the Kinova Gen3 arm attached to HyQReal, see Fig. \ref{fig:Workspace}. This approach allows the robot to perform selective, plant-by-plant task automation within the vineyard. The navigation pipeline uses an RGB-D camera instead of a more expensive, intensive sensor like LiDAR to make the system more cost-effective. However, an RGB-D camera produces erroneous camera detections that the architecture must overcome to be effective in a broader array of conditions. This paper proposes a reliable navigation architecture that generates precise waypoints based on RGB-D computer vision and higher-level control algorithms for quadruped robots.

The proposed navigation stack moves autonomously through vineyards, moving to precise waypoints autonomously generated from camera input. The navigation stack uses computer vision to detect the grapevine trunks and a filter using a rolling average of the detections to eliminate noisy input. It then generates waypoints to approach based on the detected grapevine trunks. We implemented instance segmentation for grapevine trunks using an RGB-D sensor. Detections of the grapevine trunks are made using a Mask-RCNN trained off a manually annotated dataset with over 500 images. Combining the higher-level control with the grapevine detections makes the basis for the VINUM navigation stack. We performed a series of experiments with HyQReal, and our approach achieved a mean of 21.5cm and a standard deviation of 17.6cm of distance from the desired position. Since the kinematic workspace of the Kinova Gen3 arm is larger than the distance between the first and last spur of a single grapevine plant (typically 70cm for a spur-pruned vine),\ the achieved results are acceptable for this automated task.
 
 The contributions of this proposed autonomous navigation architecture are:
 \begin{itemize}
     \item Higher-level control algorithms for precise robot placement, ideal for a robotic workspace to perform selective, plant-by-plant task automation in the vineyard row.
     \item Algorithms for accurately generating navigation waypoints from RGB-D sensors during runtime, while navigating down the vineyard row.
     \item A hand-annotated instance segmentation \href{https://universe.roboflow.com/vinum/potted_grapevines/dataset/2}{dataset} of grapevine trunks, used for the training of our Mask-RCNN.
 \end{itemize}

%% file: sections/state_of_the_art.tex
The development of autonomous robots and vehicles that can navigate vineyards has significantly advanced in the field \cite{vineyard_slam}.. While many of these robots are wheeled and move continuously throughout the vineyard row, some non-continuous models require manually added waypoints. Currently, systems can generate waypoints for autonomous navigation, but these typically rely on initial aerial maps of the vineyard, which limits their effectiveness in adapting to changing conditions at runtime.
The EU Project BACCHUS robot is a wheeled vehicle under development to harvest grapes and care for vineyards. The BACCHUS robot uses semantic segmentation of vineyard trunks for accurate robot motion estimation and consistent metric maps \cite{loop_closure}. Our proposed navigation architecture takes the same segmentation approach, but it is used to identify discrete positions for the robot to walk to instead. Previous vineyard navigation has described moving down each row, using a laser sensor, until there are no more grapevines in a row \cite{inproceedings}. Other navigation stacks also move autonomously down vineyard rows, using laser scanners for perception \cite{rob_farmer}\cite{Nehme2021}.
The EU Project CANOPIES aims to develop a human-robot collaborative paradigm for harvesting and pruning in vineyards\footnote[1]{www.canopies-project.eu}. It is a wheeled robot directed by human workers to perform precision agriculture tasks. A similar autonomous over-the-vineyard row robot is the ViTiBOT Bakus which is used to improve vineyard help by removing herbicides and using precision spraying. This solution does not include stopping at each grapevine. YANMAR's autonomous over-the-row robot, YV01, does a similar task that autonomously sprays vineyard rows without stopping at a specific grapevine\footnote[2]{https://www.yanmar.com/eu/campaign/2021/10/vineyard/}.
Another wheeled robot for precision agriculture is the Agri.q02, which is meant to work in unstructured environments in collaboration with a UAV \cite{agriq}. A navigation stack was created for the wheeled Ackerman Vehicles in precision farming, path planning from pose to pose \cite{ackerman}. There was autonomous navigation outlined in the Echord++ GRAPE experiment, which maps a vineyard that uses a wheeled robot and moves to locations on the map to perform tasks \cite{echord_grape}. Other proposed navigation architectures use deep learning to understand a camera’s depth sensors. However, their movement continues down the vineyard row \cite{deep_learning}. 
Other work has been done with deep learning for navigation, where the robot learns to navigate to the end of a vineyard row, but it is also in continuous motion \cite{Martini_2022}. In the case where a robot would need to stop at specific locations to prune, the navigation of the robot had the locations inserted manually into its path \cite{bumblebee}. Other robots are built for the winter pruning of grapevines but do not have autonomous navigation to reach the grapevines themselves \cite{pruning_grapevines}. 
These autonomous robots are all-wheeled; most do not have to stop at precise locations in the vineyard, and those that stop at exact locations have manually inserted those locations ahead of time. 
Work has been done in generating waypoints for vineyards for navigation purposes. However, these works take an initial aerial overview and derive a list of poses from the initial map {\cite{MAZZIA2021106091}\cite{contra_cluster}\cite{9784767}}. 

Our approach generates waypoints in real time as the robot moves throughout the row without requiring initial drone mapping. This paper's proposed quadruped navigation architecture introduces a way to automate waypoint generation at runtime for navigating to every plant in the vineyard for precision agricultural tasks.

%% file: sections/2_navigation_architecture.tex
The navigation architecture is a combination of higher-level control and grapevine detection.  The higher level control will decide on the robot’s movement path through a vineyard row based on the waypoints generated from the grapevine trunks detected.  The grapevine detection is an instance segmentation of grapevine trunks  for HyQReal to approach for pruning. The instance segmentation was done by training a Mask-RCNN from Detectron2 \cite{wu2019detectron2}.

\subsection{Higher Level Control}
\begin{figure}[h]
\centering
\includegraphics[width=0.40\textwidth]{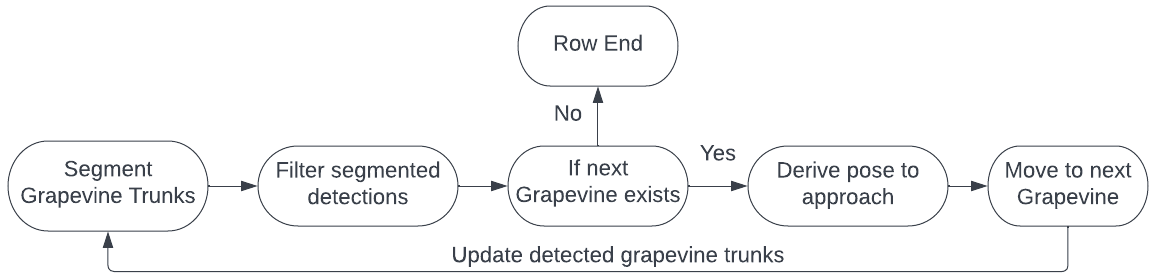}
\caption{Navigation Flow.}
\label{fig:Nav}
\end{figure}

The higher level control is a state machine for the robot to move throughout a vineyard row, as illustrated in Fig. \ref{fig:Nav}. The process begins with an initial search to find the starting lines for both sides of the row. 
The initial detections get sent through a filter to find each detection's rolling averages. From the filtered detection points, the control will see the lines on which the vineyard rows begin. After determining the vector of the row, the waypoints are generated to approach.

 During the initial search, the Mask-RCNN model classifies random background objects as grapevines and gets multiple detections of the same grapevine. Each new grapevine detection is filtered to sort through the noisy detections. Every time a grapevine trunk was detected, a ROS subscriber would listen to the detection and determine if the identified grapevine was newly detected or detection of a  grapevine had already been found. These detections are processed in the Algorithm \ref{alg:antiClus}.
 
The filter performs clustering on a list of "grapevine poses". The clustering is done based on the x and y positions of the poses. If any new grapevine trunk detection was within a set radius of a previously detected trunk, the new trunk detection was combined with the established trunk by taking a rolling average of the detections. The calculations are seen in the (\ref{eqn:antiClus}), where $p_{1x/y}$ is the existing grapevine detection's respective x and y coordinate, $p_{2x/2y}$ is the new grapevine detection's x, and y coordinates and $a$ is the number of times $p_{1}$ has been averaged to that point. The more detections a given point has, the more reliable the point's location.

\begin{equation}
    \label{eqn:antiClus}
    p_{1x} = \frac{p_{1x}a + p_{2x}}{a + 1} \\ 
    p_{1y} = \frac{p_{1y}a + p_{2y}}{a + 1} \\
\end{equation}

\begin{algorithm}
\caption{Cluster Detections}
\label{alg:antiClus}
 \hspace*{\algorithmicindent} \textbf{Input: Grapevine Pose $gp$} \\
 \hspace*{\algorithmicindent} \textbf{Output: Poses of clustered grapevine trunks $cgt$}
 \begin{algorithmic}
\State averaged Grapevine poses $agp$ tupled with times averaged $ta$ $(agp, ta) \gets empty$
 \If{$agp$ = 0}
 \State $agp \gets gp$
 \State $ta \gets 1$
 \Else
 \For{every pose tuple $pt$ in $agpt$}
 \State get x and y bottom and top around the grapevine's location 
 \State $xb, xt, yb, yt$
 \If{$xb \leq tpx \leq xt \And yx \leq tpy \leq yt$}
 \State pose tuple x $ptx$ = ($ptx$ * $ta$ + $gpx$) / ($ta$ + 1.0)
 \State pose tuple y $pty$ = ($pty$ * $ta$ + $gpy$) / ($ta$ + 1.0)
 \State $ta$ + 1
 \EndIf
 \EndFor
 \If{$gp$ is not in $atp$}
 \State $agp \gets gp$
 \EndIf
 \State $cgt \gets $ all grapevine poses $gp$ in $agp$ where $gp_i$'s  $ga_i > 2$ times
 \EndIf
 \end{algorithmic}

\end{algorithm}

After determining the grapevines, the robot has to approach in parallel to the grapevines to prune correctly. To find the correct destination point, initially, the robot determines the orientation of the approach by calculating the vector of the vineyards in a row. This is derived from a list of points found in the initial search. Each possible row line could be found using the determinant of the grapevine points detected from the list of points, as shown in (\ref{eqn:det}). If the determinant is 0, the list of points is a line. In the real-world trials, however, the determinant of the lines found was never 0. $\epsilon$ was used to set the determinant size of a line to allow for more leniency in finding lines.

\begin{gather}
\label{eqn:det}
 \det\begin{bmatrix} 
 X_{i} & X_{j} \\
 Y_{i} & Y_{j} 
 \end{bmatrix} \leq \epsilon
\end{gather}

\begin{figure}[h]
\centering
\includegraphics[height=4.0cm, width=0.42\textwidth]{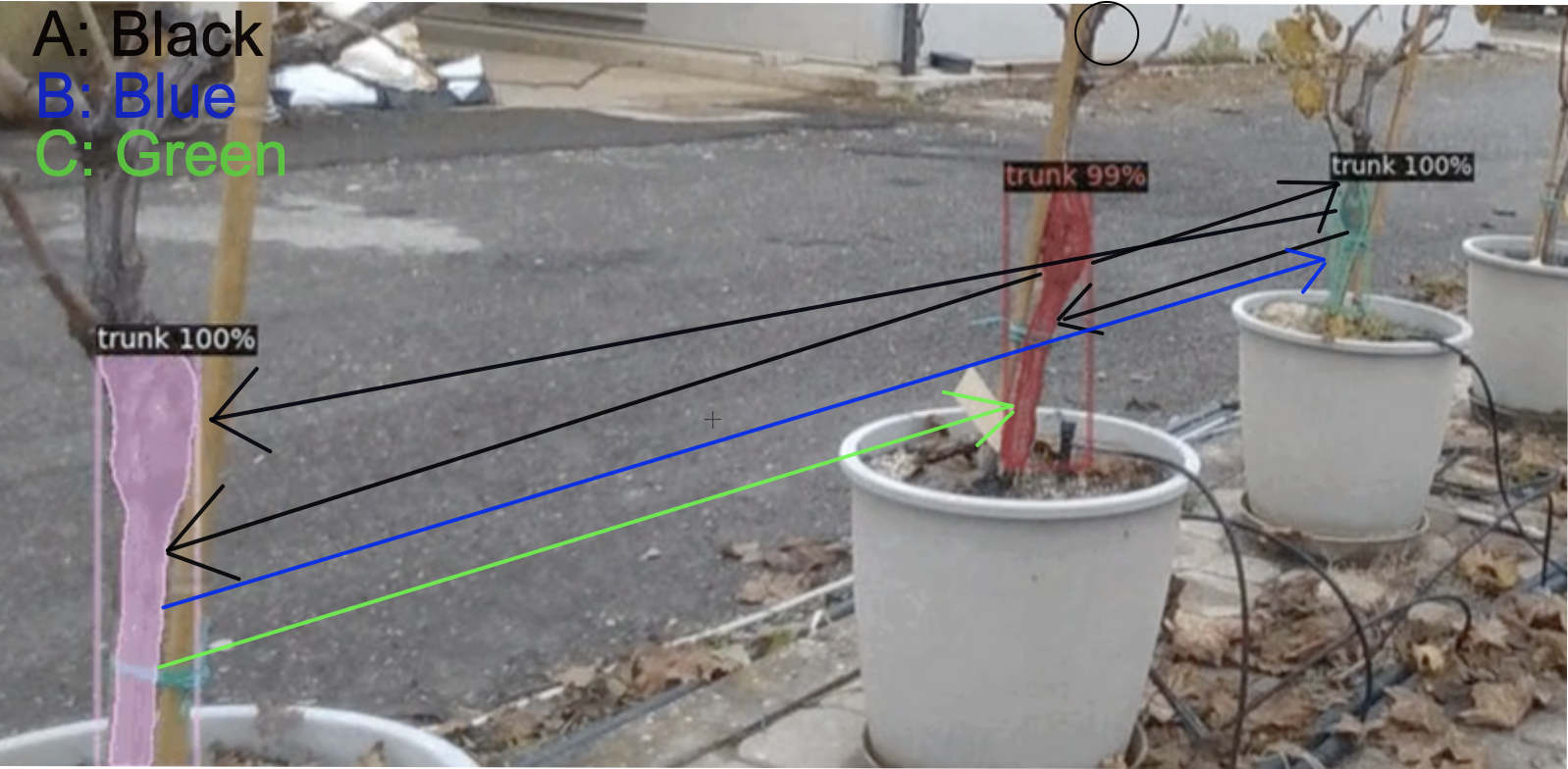}
\captionsetup{justification=centering}
\caption{Segmented Grapevine trunk image showing three detected grapevine trunks which are part of the initial line.}
\label{fig:sub}
\end{figure}

The initial vector is determined by taking the closest grapevine pose found $gp$, taking the subset of the found lines which only have the ones with the closest pose $A$, then determining the subset of $A$ where $gp$ is the first pose $B$ in the found line. Then take the line with the shortest distance from $B$. The assumption is that after the initial search, a vineyard row found would be a shorter line than the line found between two rows. See Fig. \ref{fig:sub}.

Let $ G$ denote the set of all grapevine trunk poses. The subset of lines found with the closest trunk pose to $gp$ is denoted by $A = {L_i | L_i \subseteq G, L_i \text{ has the closest pose to } gp}$. The subset of lines in $A$, where $gp$ is the first pose, is denoted by $B = {L_j | L_j \in A, gp = \text{ first pose in } L_j}$. The line with the shortest distance from $B$ is denoted by $C = \min_{L_j \in B} \text{ distance between } L_j \text{ and } B$. $ C$ determines the initial vector.

After the initial row vector for the robot is determined, the robot must approach at a parallel offset at a desired distance $d$ to maximize the pruning arm's robotic workspace. A new transformation is created with the grapevine trunk at the center to ensure a parallel approach. This transformation is calculated using the angle of the row's vector to the world frame. The Euler angles are converted into a quaternion, and the grapevine trunk's $x$, $y$ world positions are set to the center of the trunk frame. Finally, the destination pose is determined by adding an offset of $d$ in the lateral direction of the grapevine.

The higher level control updates the vector for possible deviances of grapevines as the robot moves along the row.  The robot then approaches the grapevines in parallel at a desired distance that depends on the robot's size and the arm's workspace. 

After the robot has reached the determined location in the vineyard, the control stack removes that grapevine from the list of vines to approach. Next, the control will choose the closest grapevine to the robot as its next target. This method is continued until no more grapevines are identified in a row. 

\subsection{Grapevine Identification}
Instance segmentation is a computer vision task that involves identifying and segmenting individual objects in an image or video. In the case mentioned, a Mask R-CNN is used, for instance, segmentation, to detect grapevine trunks in a vineyard. The Mask R-CNN is a two-stage deep learning model that generates region proposals and classifies and segments the regions of interest.

\begin{figure}[h]
\centering
\captionsetup{justification=centering}
\includegraphics[height=4.5cm, width=0.40\textwidth]{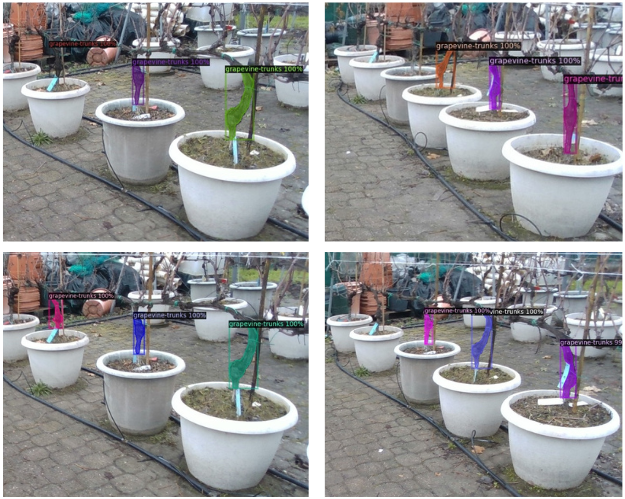}
\caption{Result of the image segmentation to detect grapevine trunks. (4 examples).}
\label{fig:trunk_det}
\end{figure}

The training of the Mask R-CNN was done using Detectron2, an open-source platform for object detection research. 500+ hand-annotated images of potted grapevines were created and used for this training\footnote{https://universe.roboflow.com/vinum/potted$\textunderscore$grapevines/dataset/2}. These annotated images helped the neural network to learn the characteristics of grapevine trunks and improved its ability to detect them in new, unseen images.

An aligned depth image was used to determine the depth of the detection; the depth image provided the necessary information to calculate the distance between the grapevine trunks and HyQReal. This information was used to find the locations of the grapevine trunks relative to the quadruped. However, depth readings were not always reliable, so to account for depth errors from the sensor, we created an algorithm to derive the closest possible valid depth value, seen in Algorithm \ref{alg:depth}.

Algorithm \ref{alg:depth} takes in two inputs: $x$ and $y$, and computes the corresponding 3D coordinates $(realX, realY, realZ)$ of a point in a given depth image $depth$. The algorithm first sets a variable $npixels$ to 2 and computes the depth image's height $h$ and width $w$. It then enters a loop, where it gradually expands a square around the point of interest, defined by the input $x$ and $y$, until the mean depth value in that square is found, stored in the variable $realZ$. The loop continues until $realZ$ is no longer $NaN$ or the square has reached the boundaries of the depth image.

Once $realZ$ is found, the function computes the corresponding 3D coordinates using the intrinsic parameters of the camera, stored in the $camInfoMsg.K$ matrix and returns the 3D coordinates as $realX, realY, realZ$.

\begin{algorithm}
\caption{Reliable Depth Value}
\label{alg:depth}
\textbf{Input:} 2D image coordinates $(x, y)$, depth image $depth$, camera information message $camInfoMsg$ \\
\textbf{Output:} 3D real world coordinates $(realX, realY, realZ)$
\begin{algorithmic}
\State $nPixels \gets 2$
\State $h, w \gets depth.shape$
\State $realZ \gets \text{NaN}$
\State $i \gets 1$
\State $minY, maxY, minX, maxX \gets y, y, x, x$
\While{$realZ$ is NaN and not ((minY == 0) and (minX == 0) and (maxY == h - 1) and (maxX == w - 1))}
\State $minY \gets \text{clip}(y - i * nPixels, 0, h - 1.0)$
\State $maxY \gets \text{clip}(y + i * nPixels, 0, h - 1.0)$
\State $minX \gets \text{clip}(x - i * nPixels, 0, w - 1.0)$
\State $maxX \gets \text{clip}(x + i * nPixels, 0, w - 1.0)$
\State $subImg \gets depth[minY:maxY + 1.0, minX:maxX + 1.0]$
\State $realZ \gets \text{nanmean}(\text{asarray}(subImg))$
\State $i \gets i + 1$
\EndWhile
\State $cx, cy, fx, fy \gets camInfoMsg.K$
\State $realX \gets (x - cx) * realZ / fx$
\State $realY \gets (y - cy) * realZ / fy$
\State \textbf{return} $(realX, realY, realZ)$
\end{algorithmic}
\end{algorithm}

A Mask-RCNN previously trained in Detectron2 framework was used \cite{DBLP:journals/corr/abs-2109-07247}. The results of the trained network can be seen in Fig. \ref{fig:trunk_det}.

%% file: sections/4_experiment.tex
\begin{figure}[h]
\centering
\includegraphics[height=4.0cm,width=0.44\textwidth]{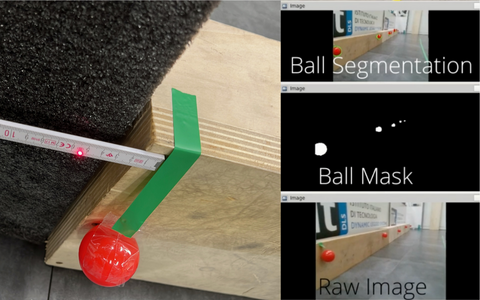} 
\captionsetup{justification=centering}
\caption{Measurement of Aliengo's arrival at a position using a laser pointer mounted onto the robot and the Image Segmentation of Red Balls.}
\label{fig:exp_measure}
\end{figure}
\begin{figure}[h]
\centering
\includegraphics[height=4.0cm, width=0.40\textwidth]{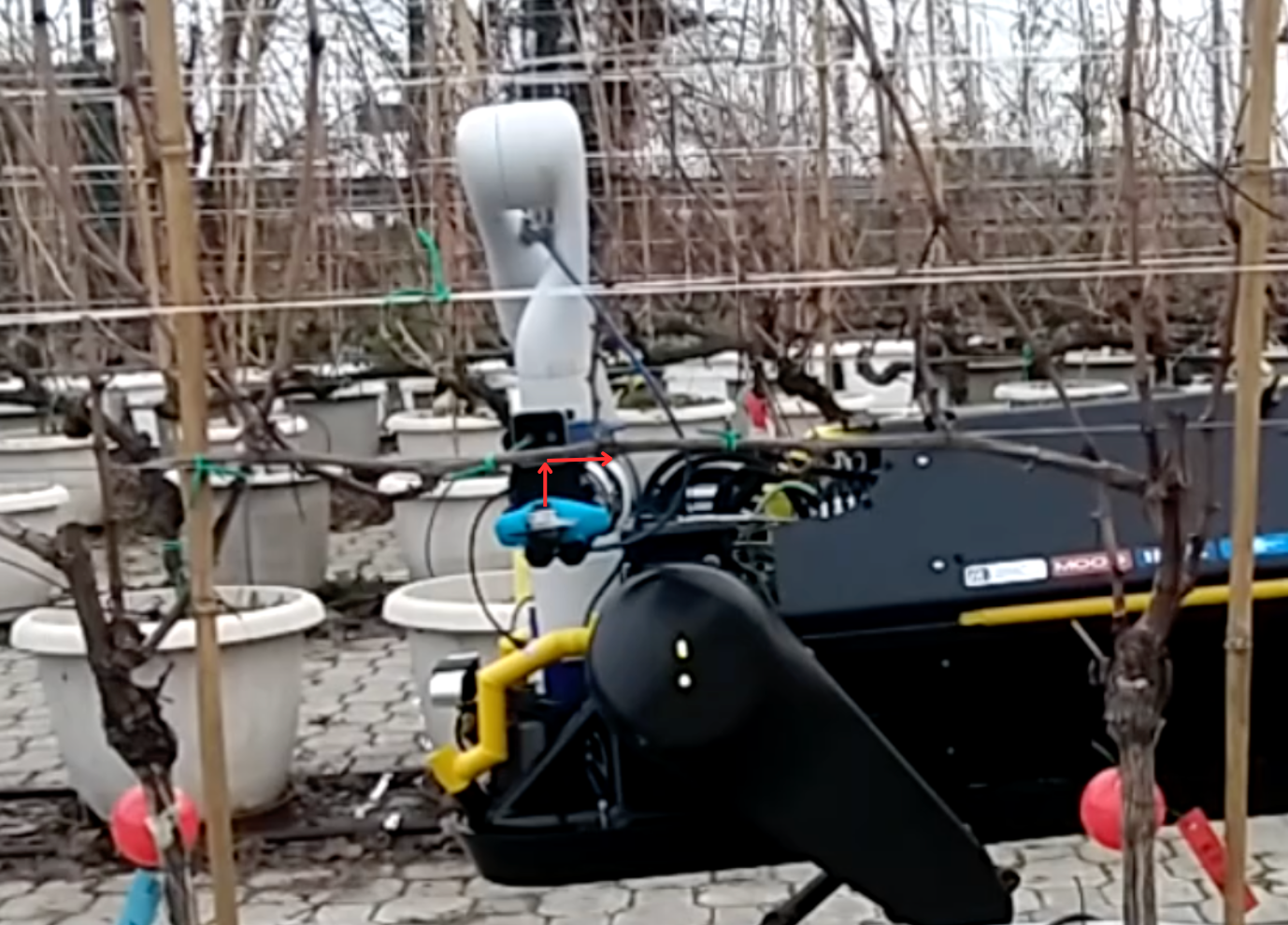} 
\captionsetup{justification=centering}
\caption{Measurement of the shear to the center of the cordon.}
\label{fig:arm_wkspace}
\end{figure}

These experiments aim to determine the precision of the waypoints generated for the quadruped robot to move to. They aim to align the robot's geometric center with the grapevine trunk; this way, an arm mounted on the front of HyQReal is in the center of the grapevine's cordon and thus optimizes the workspace of the arm for single-plant operations.
\begin{figure}[h]
\centering
\includegraphics[height=1.5cm, width=0.40\textwidth]{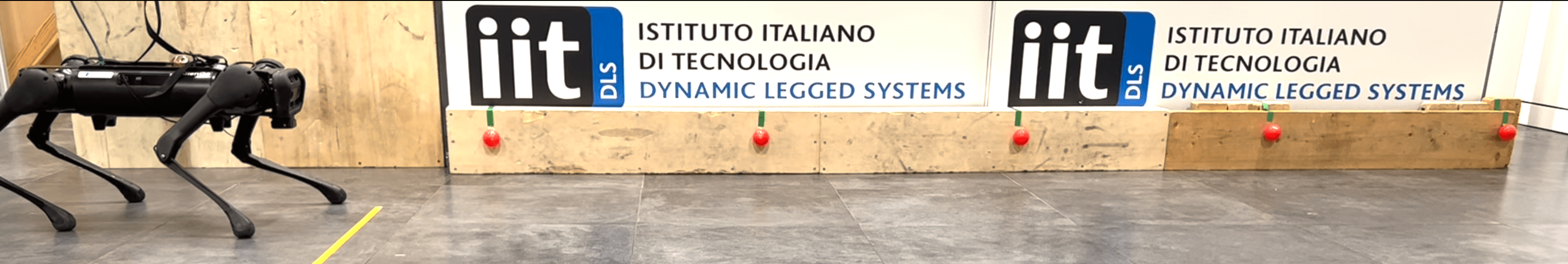}
\captionsetup{justification=centering}
\caption{Experimental setup in the lab with the quadruped robot Aliengo.}
\label{fig:exp_setup}
\end{figure}
Two rounds of testing were conducted; initially, the higher level control was tested on Aliengo, then the architecture was moved onto HyQReal. The higher level control was tested in a lab using Unitree's Aliengo robot. Aliengo was used to simplify experiments since it is 21kg and 61cm long, as compared to the 140kg and 1.3m length of HyQReal. Aliengo is equipped with Intel's Realsense D435 RGB-D camera. Red balls for segmenting were used to test in the lab instead of the grapevine trunks; see Fig. \ref{fig:exp_measure}. The red balls are spaced out at about 80cm from each other, the approximate distance that grapevines are from each other. The setup of the experiment can be seen in Fig. \ref{fig:exp_setup}. How the precision of the robot approaching a position was measured shown in Fig. \ref{fig:exp_measure}.
The tests were then repeated with HyQReal in the lab, equipped with the ZED depth camera. HyQReal was also tested in the vineyard at the University Cattolica of Piacenza with potted vines. 
During the tests, the robot initially searches the area using its RGB-D camera to segment the red balls. After the robot finds the row of red balls, it approaches the first position in the row. After the robot arrives at the initial pose, it pauses for an automated task and updates its detections. This process repeats until the row is finished and then stops.
\begin{figure}[h]
\centering
\includegraphics[height=4.0cm, width=0.40\textwidth]{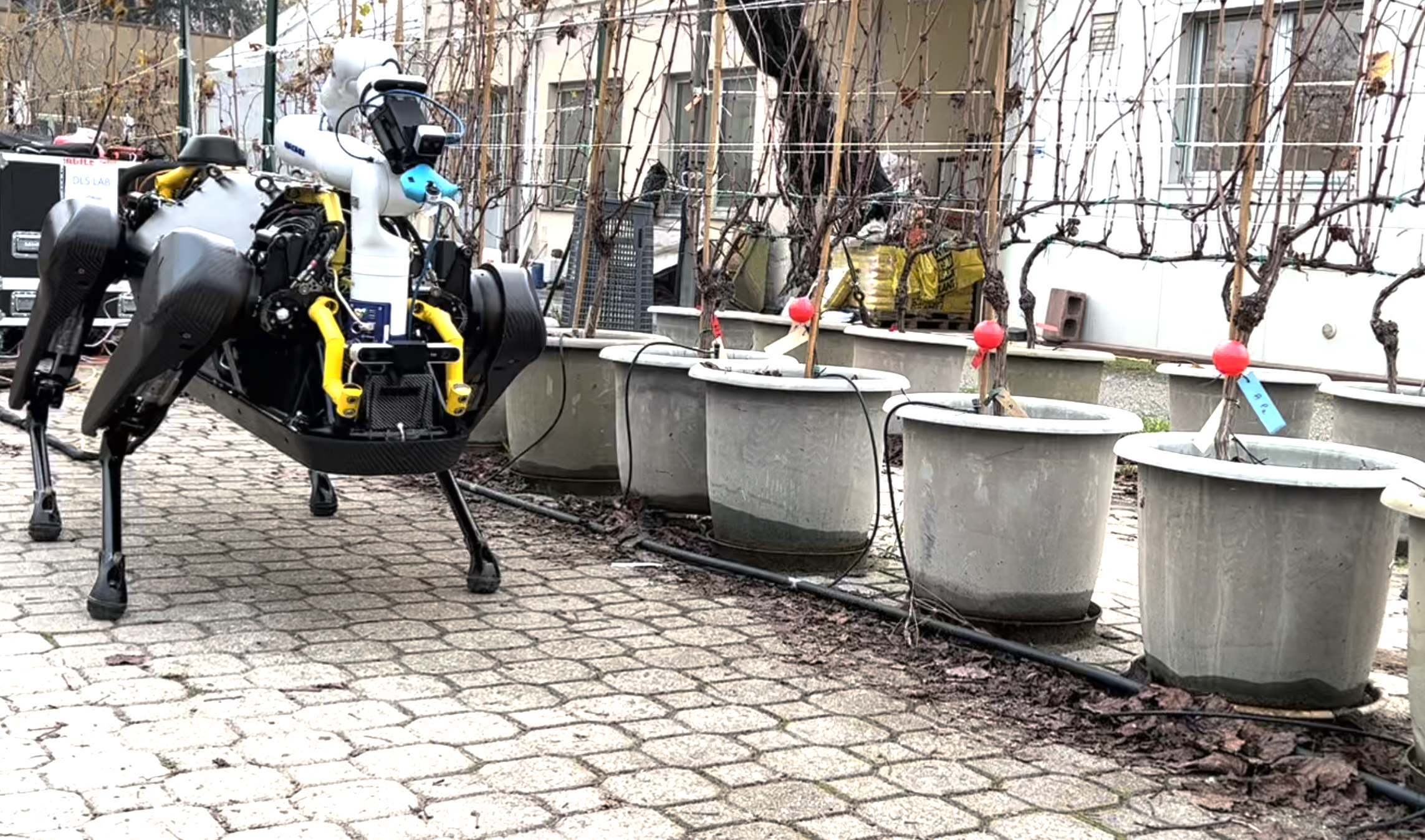}
\caption{Piacenza experiment setup with HyQReal.}
\label{fig:piacenza}
\end{figure}
With Aliengo, ten trials were conducted with five balls. A laser pointer mounted on the robot's torso was used to show the point that Aliengo's center of mass reached to measure the error between the achieved destination point and the red ball. With HyQReal, five trials were conducted with three red balls; see Fig. \ref{fig:piacenza}. The error was measured from the base of the robotic arm mount on HyQReal to the center of the grapevine's cordon; see Fig. \ref{fig:arm_wkspace}.
Aliengo’s error of reaching the destination point has a mean of 3.4cm and a standard deviation of 2.2cm. 
HyQReal’s error of reaching the destination point has a mean of 21.5cm and a standard deviation of 17.6cm.
The tests showed a reliable higher-level control with repeatable navigation. However, the results are significantly predicated on an accurate initial search. The grapevine segmentation was initially tested in Piacenza. However, the model trained had difficulty generalizing different points of view needed for identifying the grapevines, so we used red balls in the interim to test the navigation architecture. In cases where HyQReal did not accurately identify a grapevine, it would skip over. In tests in Piacenza, there was fog one morning, and the ZED camera could not make reliable detections, resulting in skipped grapevines. However, when the grapevine search went well, the robot reliably approached each grapevine.

%% file: sections/conclusion.tex
This paper presented a method of computer-vision-based navigation in vineyards for quadruped robots. This method will allow for precise waypoint generation to perform selective task automation.  Our approach with HyQReal achieved a mean of 21.5cm and a standard deviation of 17.6cm of distance from the desired position, which is sufficient for an automated task in a vineyard.

The system’s architecture works reliably, generating and arriving at precise waypoints that maximize the attached robotic arm’s workspace.  However, there could be more accuracy in grapevine detection as the RGB-D camera’s initial search needs to be trained off more data for more robust detections.  

The future steps for this architecture are to improve the dataset created to train a more robust Mask-RCNN and then the full integration of the navigation combined with the pruning arm for VINUM autonomous winter pruning of grapevines.

%% file: sections/7_acknowledgments.tex
Thanks to the contributions of Professors Matteo Gatti and Stefano Poni from the Universita Cattolica del Sacro Cuore for their important inputs regarding the experiments. Also, thanks to Chundri Boelens for helping set up HyQReal and Lorenzo Amatucci for the configuration of the robot's controller.